\documentclass[11pt,a4paper]{article}
\usepackage[hyperref]{emnlp2020}

\usepackage{natbib}
\usepackage{latexsym}
\usepackage{times}

\usepackage{microtype}

\usepackage{amsmath}
\usepackage{amsthm}
\usepackage{booktabs}
\usepackage{algorithm}
\usepackage{algorithmic}
\usepackage{booktabs}
\usepackage{multirow}
\usepackage{graphics}
\usepackage[normalem]{ulem}
\usepackage{todonotes}
\usepackage{svg}

\usepackage{xcolor,colortbl}

\definecolor{Gray}{gray}{0.85}
\definecolor{LightCyan}{rgb}{0.88,1,1}
\newcolumntype{a}{>{\columncolor{Gray}}c}

\usepackage{subfig}

\usepackage{caption}

\aclfinalcopy 
\def\aclpaperid{2145} 

\title{Corpora Evaluation and System Bias Detection \\in Multi-document Summarization}

\author{
    Alvin Dey$^{\dagger}$\thanks{Equal contribution; listed alphabetically.} \quad
	Tanya Chowdhury$^{\diamondsuit^\dagger}$\footnotemark[1]\quad
	Yash Kumar Atri$^{\dagger}$\footnotemark[1] \quad 
	 Tanmoy Chakraborty$^{\dagger}$ \\
	$^{\diamondsuit}$University of Massachusetts, Amherst, MA, USA \\
	$^{\dagger}$IIIT-Delhi, New Delhi, India \\
	{\tt tchowdhury@cs.umass.edu }\\
	{\tt \{alvin18066,yashk,tanmoy\}@iiitd.ac.in}
}

\date{}

\begin{document}

\maketitle
\begin{abstract}
  Multi-document summarization (MDS) is the task of reflecting key points from any set of documents into a concise text paragraph. In the past, it has been used to aggregate news, tweets, product reviews, etc. from various sources.  
  Owing to {\em no standard definition of the task}, we encounter a plethora of datasets with varying levels of overlap and conflict between participating documents. There is also {\em no standard regarding what constitutes summary information} in MDS.
  Adding to the challenge is the fact that new systems report results on a set of chosen datasets, which might not correlate with their performance on the other datasets.
  In this paper, we study this heterogeneous task with the help of a few widely used MDS corpora and a suite of state-of-the-art models. We make an attempt to quantify the quality of  summarization corpus and prescribe a list of points to consider while proposing a new MDS corpus. Next, we analyze the reason behind the absence of an MDS system which achieves superior performance across {\em all} corpora. We then observe the extent to which system metrics are influenced, and bias is propagated due to corpus properties. The scripts to reproduce the experiments in this work are available at \small{\color{blue}\url{https://github.com/LCS2-IIITD/summarization_bias.git}}.

\end{abstract}

\section{Introduction}
Multi-document summarization (MDS) deals with compressing more than one document into a textual summary. 
It has a wide range of applications  --  gaining insights from tweets related to similar hashtags,  understanding product features amongst e-commerce reviews, summarizing live blogs related to an ongoing match,  etc.   
Most studies on MDS were performed during the DUC\footnote{https://duc.nist.gov/} and TAC\footnote{http://tac.nist.gov
} challenges starting in the early 2000s. Each version of the challenges released a new dataset. Most of the MDS systems submitted to these challenges were unsupervised and extractive in nature. Gradually, the data released in these challenges became the {\em de facto} for MDS. These datasets were manually curated and had less than a hundred instances each. The recent development of deep neural architecture has led to a significant increase in the number of supervised document summarization systems. Large labeled corpora which are mostly crowd-sourced have been introduced to meet the training requirements of the supervised systems. However, the crowd-sourced corpora widely differ in quality based on factors like genre, size of the community, presence of moderation in the community, etc. This is further aggravated by the complexity of the task, the hardness of accumulating labeled data, or more so in the definition of what constitutes a multi-document summary.

Recently, a few large datasets for MDS have been introduced \citep{Multinews2019,CQASumm2019}. However, there has been no study to measure the relative complexity of these datasets. We observe that existing  MDS systems behave differently on different corpora. For example, a system achieving state-of-the-art performance on one corpus fails to achieve reasonable performance on another. Although the ROUGE points of MDS systems are increasing day-by-day, manual inspection reveals an increased presence of bias in generated summaries. New systems are being introduced and evaluated on a few selected corpora, leading to difficulty in understanding whether the bias is introduced by the system or it is present in the corpus used for training. 


Our research questions are as follows:
\newline
     \noindent {\textbf{Q1.} How should one model the quality of a MDS corpus as a function of its intrinsic properties?}
      \newline
      {\textbf{Q2.} Why do the ROUGE-based ranks of different MDS systems differ across different corpora? How should an MDS system which intends to achieve high ROUGE scores across {\em all} corpora, look like?  }
     \newline 
      {\textbf{Q3.} Why do systems show bias on different metrics, and which other system and corpus attributes are the reason behind it?}
      \newline
      {\textbf{Q4.} Is the task of MDS almost solved, or is there still scope for improvement? }
      \newline
We study five MDS corpora --  DUC \citep{duc}, TAC \citep{tac}, Opinosis \citep{Opiniosis2010}, Multinews \citep{Multinews2019}, and CQASumm \citep{chowdhury2019cqasumm}. We consider eight popular summarization systems -- LexRank \citep{Erkan_2004}, TextRank \citep{textrank2004}, MMR \citep{10.1145/3130348.3130369}, ICSISumm \citep{favre2008tac}, PG \citep{see2017get}, PG-MMR \citep{lebanoff-song-liu:2018}, Hi-Map \citep{Multinews2019}, and CopyTransformer \citep{gehrmann2018bottom}.

Our major contributions are four-fold:\\
$\bullet$ We propose a suite of metrics to model the quality of an MDS corpus in terms of -- Abstractness, Inter Document Similarity (IDS), Redundancy, Pyramid Score, Layout Bias and Inverse-Pyramid Score.\\
$\bullet$ We develop an interactive web portal for imminent corpora to be uploaded and evaluated based on our proposed metrics.\\
$\bullet$ We explore different biases that the MDS systems exhibit over different corpora and provide insight into properties that a universal MDS system should display to achieve reasonable performance on all types of corpora. \\
$\bullet$ We look into metrics to capture bias shown by MDS systems and explore the extent to which corpus properties influence them.

{\bf To the best of our knowledge, the current study is the first of its kind.}
\section{Background and Proposed Metrics}
Throughout the paper, we use the term \textbf{candidate documents} for the documents participating in summarization, and the term \textbf{reference} to indicate the ground-truth summary.

{\bf Oracle summary} is the extractive set of sentences selected from the candidate documents, exhibiting maximum ROUGE-N score w.r.t. the reference summary. 
It is an NP-hard problem \cite{oraclegeneration}, and approximate solutions can be found greedily or using ILP solvers.

Here, we briefly introduce a suite of corpus and system metrics proposed by us to better understand the MDS task. These metrics are further explained in detail in Supplementary.

\subsection{Corpus Metrics}
    \label{corpusmetrics}
    $\bullet$ \textbf{Abstractness}: It is defined as the percentage of non-overlapping \textit{higher order $n$-grams} between the reference summary and candidate documents. A high score highlights the presence of more distinctive phrases in reference summary. The intuition behind quantifying the number of new words is to sync with the basic human nature of paraphrasing while summarizing.\\ 
    $\bullet$ \textbf{Inter Document Similarity (IDS)}: It is an indicator of the degree of overlap between candidate documents. Inspired by the theoretical model of relevance \citep{peyrard2019simple}, we calculate IDS of a set of documents as follows:
    \begin{equation}\small
    IDS(D_i)=\frac{\sum_{D_j\in S}Relevance(D_j,D_i)}{|S|} 
    \label{eq:IDS}
    \vspace{-1mm}
    \end{equation}
    where \(D_{i}\) is the $i^{th}$ candidate document, and $S$ is the set of all documents other than \(D_{i}\). Here, $Relevance$(.,.) is defined as:
    \begin{equation}\small
    Relevance(A,B)=\sum_{\omega_i}P_A(\omega_i).\log(P_B(\omega_i))  \label{eq:REL}
    \vspace{-1mm}
    \end{equation}
    where $P_{A}(\omega_{i})$ represents the probability distribution of the $i^{th}$ semantic unit\footnote{\label{SU}An atomic piece of information} in document $A$. The further this score is from 0, the lesser inter document overlap there is in terms of semantic unit distribution. As shown in Equation \ref{eq:IDS}, the numerator calculates relevance which can be interpreted as the average surprise of observing one distribution while expecting another. This score is small if the distributions are similar i.e., \(P_A\) $\approx$ \(P_B\) from Equation \ref{eq:REL}. \\

    $\bullet$ \textbf{Pyramid Score}: We propose the metric Corpus Pyramid score to measure how well important information across documents is represented in the ground truth. As introduced by \cite{pyramid-m}, Pyramid score is a metric to evaluate system summaries w.r.t. the pool of ground-truth summaries. We instead use this metric to quantitatively analyze the ground-truth summary w.r.t. candidate documents.  The entire information set is split into {\em Summarization Content Units} (SCUs\footnote{\label{SCU}They are subsentential units based on semantic meaning}), and each SCU is assigned a weight based on the number of times it occurs in the text. A pyramid of SCUs is constructed with an SCU's weight, denoting its level, and a score is assigned to a text, based on the number of SCUs it contains. Pyramid score is defined as the ratio of a reference summary score and an optimal summary score. Higher values indicate that the reference summary covers the SCUs at the top of the pyramid better. SCUs present at the top are the ones occurring in most articles and thus can be deemed as important. 

    \noindent$\bullet$ \textbf{Redundancy}\label{redundancy}:  The \textit{amount of information} in a text can be measured as the negative of Shannon's entropy ($H$) \citep{peyrard2019simple}.  
    \begin{equation}\small
    H(D) = - \sum_{\omega_i}P_D(\omega_i).\log(P_D(\omega_i)) \label{eq:RED1}
    \vspace{-1mm}
    \end{equation}
    where \(P_{D}\) represents the probability distribution of documents $D$, and \(\omega_{i}\) represents the $i^{th}$ semantic unit\textsuperscript{\ref{SU}} in the distribution. $H(D)$ would be maximized for a uniform probability distribution when each semantic unit is present only once.
    The farther this score is from 0, the better a document is distributed over its semantic units in the distribution, hence lesser the redundancy. As evident from Equation \ref{eq:RED}, redundancy is maximized if all semantics units have equal distribution i.e., \(P(\omega_i)\) = \(P(\omega_j)\). The idea behind using redundancy is to quantify how well individual documents cover sub-topics, which might not be the core content but important nonetheless. Thus 
    \begin{equation} \small
        Redundancy(D) = H_{max} - H(D)
        \vspace{-1mm}
    \end{equation}
    Since $H_{max}$ is constant, we obtain
     \begin{equation}\small
    Redundancy(D) = \sum_{\omega_i}P_D(\omega_i).\log(P_D(\omega_i)) \label{eq:RED}
     \vspace{-1mm}
    \end{equation}
    \noindent$\bullet$ \textbf{Layout Bias}: We define \textit{Layout Bias} across a document as the degree of change in importance w.r.t. the ground-truth over the course of candidate documents. We divide the document into $k$ segments, calculate the importance of each segment w.r.t. the ground-truth by a similarity score, and average over the sentences in the segment. Positional importance of \(D_{j}\), the $j^{th}$ sentence  in the document is denoted by:
    \begin{equation}\small
    Positional Importance(D_j) =\max_{1\leq{i}\leq{n}}sim(\overrightarrow{D_j},\overrightarrow{R_i})
    \label{eq:PB}
    \vspace{-1mm}
    \end{equation}    
    where, \(\overrightarrow{R_{i}}\) is the vector representation of the $i^{th}$ sentence in the reference, \(sim\) is a similarity metric between two sentences, and $n$ is the total number of sentences in the reference summary.\\
    A lower shift indicates that while generating reference summaries, all segments have been given similar importance within any 3-fold segmented article.\\
    $\bullet$ \textbf{Inverse-Pyramid Score (Inv Pyr)}: We propose Inverse-Pyramid score to quantify the bias that  a reference summary exhibits w.r.t. its set of candidate documents. It measures the importance given to each document in the candidate set by the reference summary as:
    \vspace{-1mm}
    \begin{equation}\small
    Inv Pyr(D,S) =Var_{j}\big({D_j}\cap{S_u}\big)
    \label{eq:MDS}
    \vspace{-1mm}
    \end{equation}
    Here, \(D\) and \(S\) are the set of candidate documents for MDS and their summary respectively, \(Var\) is the variance, \(D_{j}\) and \(S_{u}\) are the sets of SCUs\textsuperscript{\ref{SCU}} in the $j^{th}$ document of the candidate set and the reference summary respectively. 

    Higher Inv Pyr scores suggest the difference in importance given to each document while generating the summary is higher. As evident from Equation \ref{eq:MDS}, Variance across the similarities is high if the similarity scores across the document-summary pairs are uneven.
    
\begin{figure*}[!tbp]
    \centering
    \scalebox{0.28}{
    \includegraphics{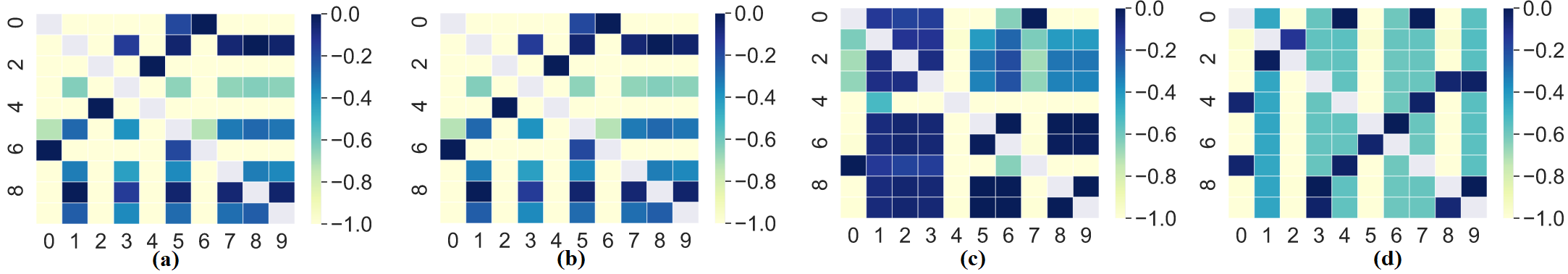}}
    \caption{Heatmap depicting the corpus metric: Inter document similarity. We explain with a single instance of (a) DUC-2004, (b) DUC-2003, (c) TAC-2008, and (d) CQASumm, and highlight inter-document overlap.}
    \label{fig:heatmap}
\end{figure*}
\begin{figure*}[!tbp]
    \centering
    \scalebox{0.25}{
    \includegraphics{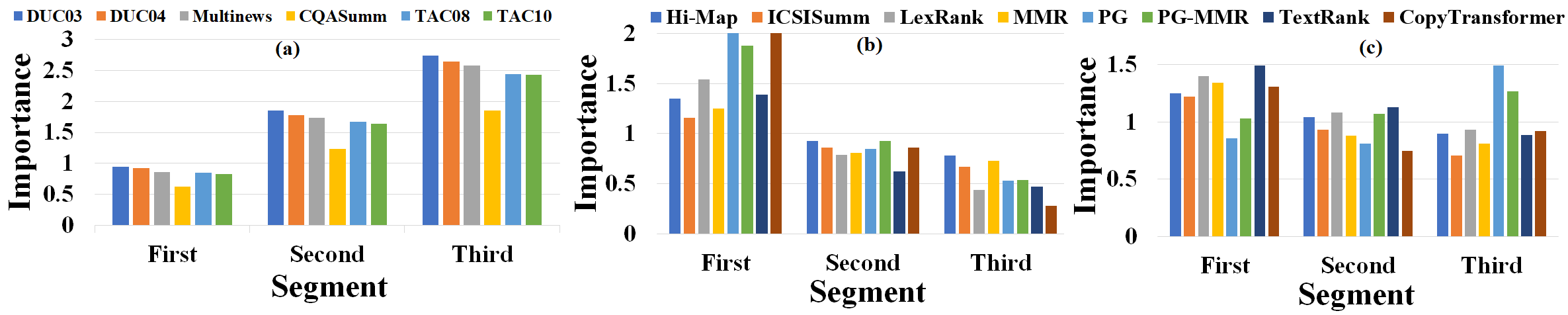}}
    \caption{(a) Layout Bias across datasets, highlighting cumulative cosine similarity (importance) values (y-axis) between segments (first, second and third) of candidate documents and the reference summary. (b) Change in layout importance across systems over source segments when divided in three uniform segments. (c) Change in layout importance across systems when candidate documents are internally shuffled and divided into three uniform segments.}
    \label{fig:position_bias}
\end{figure*}
\subsection{System Metrics}
 
    $\bullet$ \textbf{ROUGE} \label{section_rouge}  \citep{lin-2004-rouge} is a metric which computes the $n$-gram overlap recall value for the generated summary w.r.t. the reference summary.\\
    $\bullet$ \textbf{F1 Score with Oracle Summaries}:\label{section_f1} Oracle summaries reflect the extractive selection of sentences that achieve the highest ROUGE score over the candidate documents given a reference summary. Similar to ROUGE-1, this metric also combines both precision and recall between the oracle and system summaries to calculate F1 Score. It is a better indicator of the presence of non-essential $n$-grams than ROUGE as it also takes precision into account. \\ 
    $\bullet$ \textbf{System Abstractness}\label{section_abs}: Analogous to corpus abstractness, we compute the percentage of novel higher order $n$-grams in the generated summary w.r.t. the candidate documents. System abstractness is calculated using
    \begin{equation*}\small
        Coverage(D,S) = \frac{\sum_{i\in{1..n}}(D\cap S_i)}{C_n(S)}
    \end{equation*}
    where $D$ represents the set of n-grams in the candidate document, and $S$ represents the set of n-grams in the $i^{th}$ system summary. \\
    The denominator denotes the total count of n-grams in a system summary. Finally, the values of all articles is normalized to get the score for the system\\
    $\bullet$ \textbf{Layout Bias}\label{section_pos}: We propose this metric to capture which sections of the candidate documents comprise a majority of the information in the generated summary. For neural abstractive systems, we concatenate candidate documents to form one large document and feed it to the neural model. We study two variations of this metric -- The first variation involves segmenting this large document into $k$ parts and then computing the similarity of $n$-gram tokens of system summaries w.r.t. the candidate document segment. The second variation includes shuffling the candidate documents before concatenating and then computing the $n$-gram similarity with the generated summary.\\
    $\bullet$ \textbf{Inter Document Distribution (IDD)}\label{section_ids}: We propose this metric to quantify the extent of contribution of each candidate document to form the generated summary. The relevance for system summaries is calculated by,
    \begin{equation*}\small
    Relevance(A,B)=\sum_{\omega_i}P_A(\omega_i).\log(P_B(\omega_i)) 
    \end{equation*}
    where \(P_{A}\) represents the probability distribution of system summary $S$, and \(\omega_{i}\) represents the $i^{th}$ semantic unit in the distribution. 
    \begin{equation*}\small
    IDD(D_i)=\frac{\sum_{D_j\in S}Relevance(D_j,D_i)}{Cardinality(S)}
    \end{equation*} \\
    $\bullet$ \textbf{Redundancy}\label{section_red}: It measures the degree to which system summaries can cover the distribution across semantic units generated from the candidate documents.
    Redundancy for candidate documents is given by Eq.,
    \begin{equation*}
    Redundancy(D) =\sum_{\omega_i}S_D(\omega_i).\log(S_D(\omega_i))
    \end{equation*}
    where \(S_{D}\) represents the probability distribution of a system summary $D$. \(\omega_{i}\) represents the $i^{th}$ semantic unit in the distribution.

\section{Experimental Setup}
\subsection{MDS Corpora}
\noindent$\bullet$ \textbf{DUC} \citep{duc} is a news dataset built using newswire/paper documents. The 2003 (DUC-2003) and 2004 (DUC-2004) versions comprise $30$ and $50$ topics respectively with each topic having 4 manually curated reference summaries.\\
$\bullet$ \textbf{TAC} \citep{tac} is built from the AQUIANT-2 collection of newswire articles where NIST assessors select $48$ and $44$ topics for the 2008 and 2010 versions, respectively. Each topic consists of 4 summaries.\\
$\bullet$ \textbf{Opinosis} \citep{Opiniosis2010} is an accumulation of user reviews collected from various sources like TripAdvisor, Edmunds.com and Amazon. There are $51$ topics, with each topic having approximately $4$ human-written summaries.\\ 
$\bullet$ \textbf{CQASumm} \citep{chowdhury2019cqasumm} is a community question answering dataset, consisting of $100,000$ threads from the Yahoo! Answers L6 dataset. It treats each answer under a thread as a separate document and the best answer as a reference summary. \\
$\bullet$ \textbf{Multinews} \citep{Multinews2019} is a news dataset comprising news articles and human-written summaries from newser.com. It has $56,216$ topics, with summaries of $260$ words on average written by professional editors. 

\vspace{+1mm}

\subsection{MDS Systems}

To identify bias in system-generated summaries, we study a few non-neural extractive and neural abstractive summarization systems, which are extensively used for multi-document summarization.  \\
    $\bullet$ \textbf{LexRank} \citep{Erkan_2004} is a graph based algorithm that computes the importance of a sentence using the concept of eigen vector centrality in a graphical representation of text.\\
    $\bullet$ \textbf{TextRank} \citep{textrank2004} runs a modified version of PageRank \citep{ilprints361} on a weighted graph, consisting of nodes as sentences and edges as similarities between sentences.\\
    $\bullet$ \textbf{Maximal Marginal Relevance (MMR)} \citep{10.1145/3130348.3130369} is an extractive summarization system that ranks sentences based on higher relevance while considering the novelty of the sentence to reduce redundancy.\\
    $\bullet$ \textbf{ICSISumm} \citep{favre2008tac} optimizes the summary coverage by adopting a linear optimization framework. It finds a globally optimal summary using the most important concepts covered in the document.\\
    $\bullet$ \textbf{Pointer Generator (PG)} network \citep{see2017get} is a sequence-to-sequence summarization model which allows both copying words from the source by pointing or generating words from a fixed vocabulary.\\
    $\bullet$ \textbf{Pointer Generator-MMR:} PG-MMR \citep{lebanoff-song-liu:2018} uses MMR along with PG for better coverage and redundancy mitigation.\\ 
    $\bullet$ \textbf{Hi-Map:} Hierarchical MMR-Attention PG model \citep{Multinews2019} extends the work of PG and MMR. MMR scores are calculated at word level and incorporated in the attention weights for a better summary generation.\\
    $\bullet$ \textbf{Bottom-up Abstractive Summarization (CopyTransformer)} \citep{gehrmann2018bottom} uses transformer parameters proposed by \cite{DBLP:journals/corr/VaswaniSPUJGKP17}; but one of the attention heads chosen randomly acts as a copy distribution.

\begin{table}[!t]
\centering
  \caption{Values of corpus metrics: Abstractness, Redundancy (Red), Inter Document Similarity (IDS), Pyramid Score (Pyr) and Inverse-Pyramid Score (Inv).}
  \scalebox{0.64}{
\begin{tabular}{|l|r|r|r|r|r|r|r|}
\hline
\multirow{3}{*}{\textbf{Dataset}} & \multicolumn{7}{c|}{\textbf{Metric}}                                                                                                                                                                                                                                                                                                   \\ \cline{2-8} 
                                  & \multicolumn{3}{c|}{\textbf{Abstractness}}                                                                         & \multicolumn{1}{c|}{\multirow{2}{*}{\textbf{Red}}} & \multicolumn{1}{c|}{\multirow{2}{*}{\textbf{IDS}}} & \multicolumn{1}{c|}{\multirow{2}{*}{\textbf{Pyr}}} & \multicolumn{1}{c|}{\multirow{2}{*}{\textbf{Inv}}} \\ \cline{2-4}
                                  & \multicolumn{1}{c|}{\textbf{1-gram}} & \multicolumn{1}{c|}{\textbf{2-gram}} & \multicolumn{1}{c|}{\textbf{3-gram}} & \multicolumn{1}{c|}{}                              & \multicolumn{1}{c|}{}                              & \multicolumn{1}{c|}{}                              & \multicolumn{1}{c|}{}                              \\ \hline
DUC                               & 11.5                                 & 54.66                                & 79.29                                & -0.21                                              & -6.6                                               & 0.35                                               & 2.64                                               \\ \cline{1-1}
Opinosis                          & 11.5                                 & 50.36                                & 76.31                                & -0.02                                              & -5.53                                              & 0.26                                               & 2.8                                                \\ \cline{1-1}
Multinews                         & 32.28                                & 67.53                                & 80.45                                & -0.8                                               & -1.03                                              & 0.4                                                & 3.8                                                \\ \cline{1-1}
CQASumm                           & 41.41                                & 80.72                                & 88.79                                & -0.22                                              & -9.16                                              & 0.05                                               & 5.2                                                \\ \cline{1-1}
TAC                               & 9.91                                 & 50.26                                & 76.17                                & -0.19                                              & -4.43                                              & 0.32                                               & 2.9                                                \\ \hline
\end{tabular}}
\label{tab:corpus}
\end{table}
\vspace*{-\baselineskip}

\begin{table}[!tbp]
\centering
\caption{Various metrics (Met) showing ROUGE Scores (ROUGE-1, ROUGE-2), F1 Score (F1) between candidate documents and oracle summaries, Abstractness (Abs.) of abstractive systems, Redundancy (Red.) in system generated summaries, Inter Document Distribution (IDD) and Inter Document Distribution Variance (IDDV) of system summaries in dataset DUC, TAC, Opin (Opinosis), Multin (Multinews and CQAS (CQASumm).}
\scalebox{0.73}{
\begin{tabular}{|c|l|r|r|r|r|r|}
\hline
\multirow{2}{*}{\textbf{System}}                                                         & \multicolumn{1}{c|}{\multirow{2}{*}{\textbf{Met.}}} & \multicolumn{5}{c|}{\textbf{Dataset}}                                                                                                     \\ \cline{3-7} 
                                                                                         & \multicolumn{1}{c|}{}                               & \multicolumn{1}{l|}{DUC} & \multicolumn{1}{l|}{TAC} & \multicolumn{1}{l|}{Opin} & \multicolumn{1}{l|}{Multin} & \multicolumn{1}{l|}{CQAS} \\ \hline
\multirow{6}{*}{\begin{tabular}[c]{@{}c@{}}Lex\\ -Rank\end{tabular}}                     & R1                                                  & 35.56                    & 33.1                     & 33.41                     & 38.27                       & 32.22                     \\ \cline{2-2}
                                                                                         & R2                                                  & 7.87                     & 7.5                      & 9.61                      & 12.7                        & 5.84                      \\ \cline{2-2}
                                                                                         & F1                                                  & 31.34                    & 31.51                    & 31.05                     & 41.01                       & 49.71                     \\ \cline{2-2}
                                                                                         & Red.                                                & -0.136                   & -0.104                   & -0.278                    & -0.29                       & -0.364                    \\ \cline{2-2}
                                                                                         & IDD                                                 & -3.377                   & -1.87                    & -3.526                    & -2.53                       & -2.17                     \\ \cline{2-2}
                                                                                         & IDDV                                                & 0.239                    & 1.62                     & 0.221                     & 0.242                       & 1.232                     \\ \hline
\multirow{6}{*}{\begin{tabular}[c]{@{}c@{}}Text\\ -Rank\end{tabular}}                    & R1                                                  & 33.16                    & 44.98                    & 26.97                     & 38.44                       & 28.94                     \\ \cline{2-2}
                                                                                         & R2                                                  & 6.13                     & 9.28                     & 6.99                      & 13.1                        & 5.65                      \\ \cline{2-2}
                                                                                         & F1                                                  & 40.8                     & 29.69                    & 31                        & 38.44                       & 46.3                      \\ \cline{2-2}
                                                                                         & Red.                                                & -0.25                    & -1.553                   & -0.342                    & -0.208                      & -0.247                    \\ \cline{2-2}
                                                                                         & IDD                                                 & -0.196                   & -5.97                    & -2.745                    & -1.879                      & -2.137                    \\ \cline{2-2}
                                                                                         & IDDV                                                & 0.799                    & 1.48                     & 0.025                     & 0.146                       & 0.744                     \\ \hline
\multicolumn{1}{|l|}{\multirow{6}{*}{MMR}}                                               & R1                                                  & 30.14                    & 30.54                    & 30.24                     & 38.77                       & 29.33                     \\ \cline{2-2}
\multicolumn{1}{|l|}{}                                                                   & R2                                                  & 4.55                     & 4.04                     & 7.67                      & 11.98                       & 4.99                      \\ \cline{2-2}
\multicolumn{1}{|l|}{}                                                                   & F1                                                  & 30.57                    & 28.3                     & 31.8                      & 42.07                       & 45.48                     \\ \cline{2-2}
\multicolumn{1}{|l|}{}                                                                   & Red.                                                & -0.266                   & -0.068                   & -0.255                    & -0.17                       & -0.288                    \\ \cline{2-2}
\multicolumn{1}{|l|}{}                                                                   & IDD                                                 & -2.689                   & -2.135                   & -3.213                    & -1.83                       & -2.059                    \\ \cline{2-2}
\multicolumn{1}{|l|}{}                                                                   & IDDV                                                & 1.873                    & 0.231                    & 0.222                     & 0.157                       & 0.126                     \\ \hline
\multirow{6}{*}{\begin{tabular}[c]{@{}c@{}}ICSI\\ -Summ\end{tabular}}                    & R1                                                  & 37.31                    & 28.09                    & 27.63                     & 37.2                        & 28.99                     \\ \cline{2-2}
                                                                                         & R2                                                  & 9.36                     & 3.78                     & 5.32                      & 13.5                        & 4.24                      \\ \cline{2-2}
                                                                                         & F1                                                  & 24.27                    & 27.82                    & 29.83                     & 44.71                       & 50.98                     \\ \cline{2-2}
                                                                                         & Red.                                                & -0.327                   & -0.283                   & -0.328                    & -0.31                       & -0.269                    \\ \cline{2-2}
                                                                                         & IDD                                                 & -3.357                   & -1.903                   & -3.244                    & -3.14                       & -2.466                    \\ \cline{2-2}
                                                                                         & IDDV                                                & 0.694                    & 0.403                    & 1.134                     & 0.239                       & 0.242                     \\ \hline
\multirow{7}{*}{PG}                                                                      & R1                                                  & 31.43                    & 31.44                    & 19.65                     & 41.85                       & 31.09                     \\ \cline{2-2}
                                                                                         & R2                                                  & 6.03                     & 6.4                      & 1.29                      & 12.91                       & 5.52                      \\ \cline{2-2}
                                                                                         & F1                                                  & 23.08                    & 26.32                    & 16.08                     & 43.89                       & 21.85                     \\ \cline{2-2}
                                                                                         & Abs.                                                & 0.017                    & 0.01                     & 0.04                      & 0.28                        & 0.065                     \\ \cline{2-2}
                                                                                         & Red.                                                & -0.16                    & -0.2542                  & -0.188                    & -0.28                       & -0.12                     \\ \cline{2-2}
                                                                                         & IDD                                                 & -2.1                     & -1.93                    & -2.1                      & -2.103                      & -0.5                      \\ \cline{2-2}
                                                                                         & IDDV                                                & 0.248                    & 0.398                    & 0.168                     & 0.391                       & 0.391                     \\ \hline
\multirow{7}{*}{\begin{tabular}[c]{@{}c@{}}PG\\ -MMR\end{tabular}}                       & R1                                                  & 36.42                    & 40.44                    & 19.8                      & 40.55                       & 36.54                     \\ \cline{2-2}
                                                                                         & R2                                                  & 9.36                     & 14.93                    & 1.34                      & 12.36                       & 6.67                      \\ \cline{2-2}
                                                                                         & F1                                                  & 24.3                     & 26.9                     & 16.39                     & 43.93                       & 21.72                     \\ \cline{2-2}
                                                                                         & Abs.                                                & 0.019                    & 0.02                     & 0.04                      & 0.275                       & 0.069                     \\ \cline{2-2}
                                                                                         & Red.                                                & -0.17                    & -0.26                    & -0.172                    & -0.29                       & -0.142                    \\ \cline{2-2}
                                                                                         & IDD                                                 & -2.4                     & -1.87                    & -1.9                      & -1.98                       & -0.72                     \\ \cline{2-2}
                                                                                         & IDDV                                                & 0.441                    & 0.274                    & 0.192                     & 0.249                       & 0.318                     \\ \hline
\multicolumn{1}{|l|}{\multirow{7}{*}{Trans.}}                                            & R1                                                  & 28.54                    & 31.54                    & 20.46                     & 43.57                       & 30.12                     \\ \cline{2-2}
\multicolumn{1}{|l|}{}                                                                   & R2                                                  & 6.38                     & 5.9                      & 1.41                      & 14.03                       & 4.36                      \\ \cline{2-2}
\multicolumn{1}{|l|}{}                                                                   & F1                                                  & 15.72                    & 17.82                    & 16.38                     & 44.54                       & 21.35                     \\ \cline{2-2}
\multicolumn{1}{|l|}{}                                                                   & Red.                                                & -0.1771                  & -0.17                    & -0.189                    & -0.18                       & -0.273                    \\ \cline{2-2}
\multicolumn{1}{|l|}{}                                                                   & Abs.                                                & 0.09                     & 0.09                     & 0.049                     & 0.319                       & 0.092                     \\ \cline{2-2}
\multicolumn{1}{|l|}{}                                                                   & IDD                                                 & -1.9148                  & -1.8677                  & -1.589                    & -1.89                       & -2.239                    \\ \cline{2-2}
\multicolumn{1}{|l|}{}                                                                   & IDDV                                                & 0.138                    & 0.172                    & 0.249                     & 0.126                       & 1.184                     \\ \hline
\multicolumn{1}{|l|}{\multirow{7}{*}{\begin{tabular}[c]{@{}l@{}}Hi\\ -Map\end{tabular}}} & R1                                                  & 35.78                    & 29.31                    & 18.02                     & 43.47                       & 31.41                     \\ \cline{2-2}
\multicolumn{1}{|l|}{}                                                                   & R2                                                  & 8.9                      & 4.61                     & 1.46                      & 14.89                       & 4.69                      \\ \cline{2-2}
\multicolumn{1}{|l|}{}                                                                   & F1                                                  & 25.89                    & 24.3                     & 20.36                     & 42.55                       & 19.84                     \\ \cline{2-2}
\multicolumn{1}{|l|}{}                                                                   & Abs.                                                & 0.14                     & 0.147                    & 0.08                      & 0.267                       & 0.07                      \\ \cline{2-2}
\multicolumn{1}{|l|}{}                                                                   & Red.                                                & -0.1722                  & -0.2002                  & -0.16                     & -0.23                       & -0.26                     \\ \cline{2-2}
\multicolumn{1}{|l|}{}                                                                   & IDD                                                 & -1.6201                  & -1.652                   & -1.8                      & -1.788                      & -2.223                    \\ \cline{2-2}
\multicolumn{1}{|l|}{}                                                                   & IDDV                                                & 0.185                    & 0.155                    & 0.209                     & 0.209                       & 0.448                     \\ \hline
\multicolumn{1}{|l|}{\multirow{2}{*}{Highest}}                                           & R1                                                  & 94.01                    & 94.07                    & 44.53                     & 79.94                       & 64.45                     \\ \cline{2-2}
\multicolumn{1}{|l|}{}                                                                   & R2                                                  & 49.85                    & 50.17                    & 5.73                      & 42.41                       & 18.38                     \\ \hline
\end{tabular}}
\label{tab:rfa}
\end{table}

\begin{figure*}[!htbp]
    \centering
    \scalebox{0.18}{
    \includegraphics{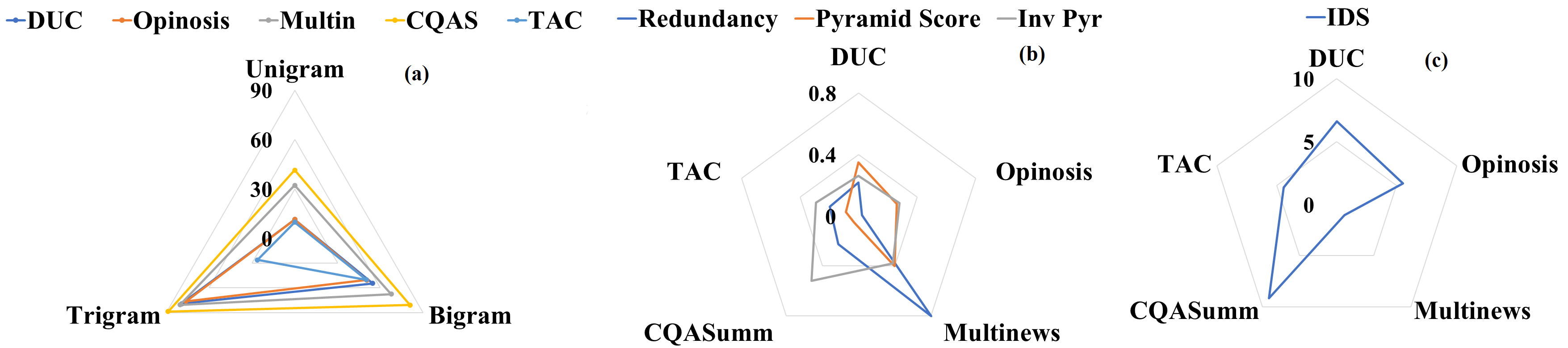}}
    \caption{(a) Abstractness across datasets. (b) Redundancy, Pyramid Score and Inverse-Pyramid Score (Inv Pyr scaled down by a factor of 10 for better visualization with other metrics) across datasets. (c) Inter Document Similarity (IDS) across datasets.}
    \label{fig:corpus_metrics}
\end{figure*}
\begin{figure*}[!htbp]
    \centering
    \scalebox{1.03}{
    \includegraphics[width=\textwidth]{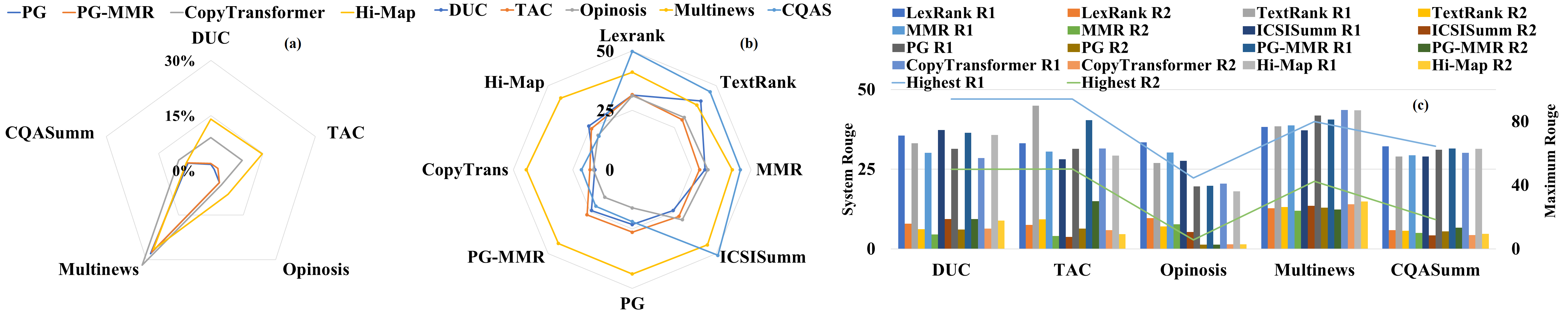}}
    \caption{(a) Level of abstractness of systems w.r.t. candidate documents and the system generated summaries. (b) F1 Score of various systems between oracle summaries and system-generated summaries. (c) ROUGE scores of various system summaries on the left axis and maximum ROUGE score over a dataset on the right axis.}
    \label{fig:system_metrics}
\end{figure*}

\section{ Inferences from Corpus Metrics} 
$\bullet$ News derived corpora show a strong layout bias where significant reference information is contained in the introductory sentences of the candidate documents (Fig. \ref{fig:position_bias}).\\ 
$\bullet$ Different MDS corpora vary in compression factors with DUC at $56.55$, TAC at $54.68$, Multinews at $8.18$ and CQASumm at $5.65$. A high compression score indicates an attempt to pack candidate documents to a shorter reference summary. \\
$\bullet$ There has been a shift in the size and abstractness of reference summaries in MDS corpora over time -- while DUC and TAC were small in size and mostly extractive ($11\%$ novel unigrams); crowd-sourced corpora like CQASumm are large enough to train neural models and highly abstractive ($41.4\%$ novel unigrams).\\
$\bullet$ Candidate documents in Opinosis, TAC and DUC feature a high degree of redundant information as compared to Opinosis and CQASumm, with instances of the former revolving around a single key entity while that of the latter tending to show more topical versatility.\\
$\bullet$ MDS corpora present a  variation in inter-document content overlap as well: while Multinews shows the highest degree of overlap, CQASumm shows the least and the rest of the corpora show moderate overlap (see Fig. \ref{fig:heatmap}).\\
$\bullet$ Pyramid Score, the metric which evaluates if the important and redundant SCUs\textsuperscript{\ref{SCU}} from the candidate documents have been elected to be part of the reference summary, shows considerably positive values for DUC, TAC and Multinews as compared to crowdsourced corpora like CQASumm (Fig. \ref{fig:corpus_metrics}.b). \\
$\bullet$ Inverse-Pyramid Score, the metric which evaluates how well SCUs\textsuperscript{\ref{SCU}} of the reference summary are distributed amongst candidate documents, also shows better performance on human-annotated corpora compared to crowd-sourced ones (Fig. \ref{fig:corpus_metrics}(b)).\\
$\bullet$ A comparison amongst corpus metrics presents a strong positive correlation between IDS and Pyramid Score (Pearson's $\rho$ = 0.8296) and a strong negative correlation between the metrics of Redundancy and IDS (Pearson's $\rho$ = -0.8454).

\begin{figure}[!t]
    \centering
    \scalebox{0.48}{
    \includegraphics[width=\textwidth]{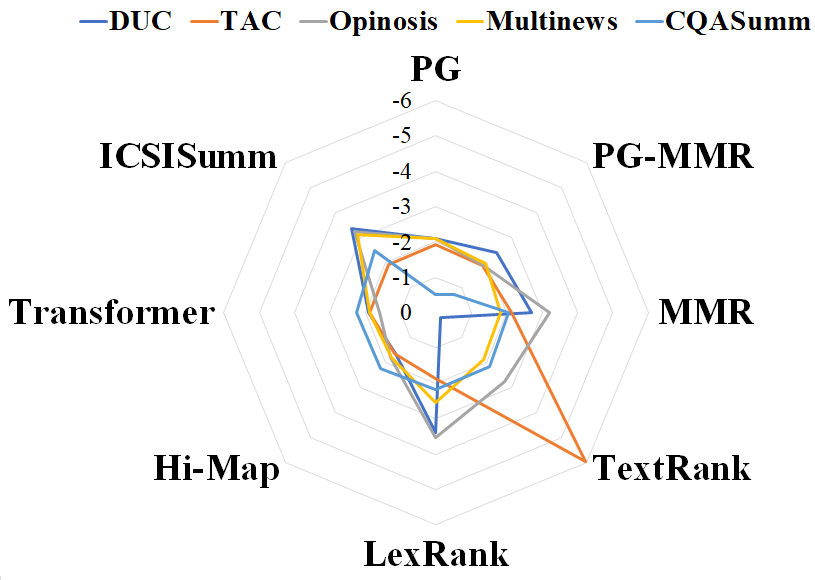}}
    \caption{Redundancy of various systems across DUC, TAC, Opinosis, Multinews and CQASumm.}
    \label{fig:appendix_system_redundancy}
\end{figure}

\section{Inferences from System Metrics}
$\bullet$ MDS systems under consideration are ranked differently in terms of ROUGE on different corpora; leading to a dilemma whether to declare a system superior to others without testing on all types of datasets (Fig. \ref{fig:system_metrics}(c)) and Table \ref{tab:rfa}).\\
$\bullet$ MDS systems under consideration outperform abstractive summarization systems by up to $10\%$ on ROUGE-1 and up to $30\%$ on F1 Scores, showing contradictory behavior in comparison to single-document summarization systems where state-of-the-art abstractive systems are known to outperform the former (Figs. \ref{fig:system_metrics}(b)-(c)).\\
$\bullet$ The best summarization system on each corpus obtains a score $39.6\%, 47.8\%, 75.02\%, 54.5\%, 49.9 \%$ of the oracle upper bound on DUC, TAC, Opinosis, Multinews and CQASumm respectively, indicating that summarization on Opinosis and Multinews is a partially solved problem, while DUC, TAC and CQASumm exhibit considerable scope for improvement (Fig. \ref{fig:system_metrics}(c)).\\
$\bullet$ Hi-Map and CopyTransformer generate more abstract summaries ($17.5\%$ and $16\%$ novel unigrams respectively) in comparison to PG and PG-MMR (Fig. \ref{fig:system_metrics}(a)).\\
$\bullet$ Averaging over systems and comparing corpora, we notice that Multinews and CQASumm achieve the highest abstractness ($27\%$ and $7\%$ respectively), which might be a result of these two corpora having the most abstract reference summaries (Fig. \ref{fig:system_metrics}(a) and (Table \ref{tab:rfa})).\\
$\bullet$ Abstractive systems exhibit a $55\%$ shift in importance between the first and the second segments of generated summaries, whereas extractive systems show an average shift of only $40\%$, implying that abstractive systems have a stronger tendency to display layout bias (Fig. \ref{fig:position_bias}(b) and Fig. \ref{fig:position_bias}(c)).\\
$\bullet$ While DUC, TAC and Opinosis summaries generated from PG trained models exhibit lower novel unigrams formation, the same for CopyTransformer and Hi-Map on DUC, TAC and Opinosis shows a higher unigram formation on average (Fig. \ref{fig:system_metrics}(a)).\\
$\bullet$ In terms of Inter Document Distribution, LexRank summary for TAC and CQASumm shows more variance across documents compared to DUC, Opinosis and Multinews. TextRank summary on DUC, TAC and CQASumm, MMR summary on DUC, and Hi-Map summary on CQASumm show higher variances as well. Systems such as PG, PG-MMR and CopyTransformer show minimal deviation in the document participation across corpora (Table \ref{tab:rfa}).\\
$\bullet$ In terms of Topic Coverage, extractive systems show better coverage than abstractive systems (Table \ref{tab:rfa}), which might be a result of extractive systems being based on sentence similarity algorithms which find important sentences, reduce redundancy and increase the spread of information from different segments of the candidate document.
(Fig. \ref{fig:appendix_system_redundancy}).

\section{Discussion on Research Questions}
\noindent\textbf{\textit{Q1. How should one model the quality of an MDS corpus as a function of its intrinsic metrics? What guidelines should be followed to propose MDS corpora for enabling a fair comparison with existing datasets?}}
The quality of an MDS corpus is a function of two independent variables: the \textit{quality of the candidate documents} and the \textit{quality of the reference summary}. 
Our findings suggest that a framework for future MDS datasets should provide scores measuring their standing w.r.t. both the above factors. The former is usually crowd-source dependent, while the latter is usually annotator dependent.
While Inter Document Similarity, Redundancy, Layout Bias and Inverse-Pyramid Score are indicators of the properties of the candidate document, metrics such as Abstractness of the reference summary and Pyramid Score are ground-truth properties. We divide the above metrics into two categories: \textit{objective} and \textit{subjective}. While all these metrics should be reported by imminent corpora proposers to enable comparisons with existing corpora and systems, we feel that the objective metrics average \textit{Pyramid Score} and \textit{Inverse-Pyramid Score} must be reported as they are strong indicators of generic corpus quality. Other subjective metrics such as IDS, Redundancy, Abstractness etc. can be modeled to optimize task-based requirements. \\


\noindent \textbf{\textit{Q2. Why do the ROUGE-based ranks of different MDS systems differ across corpora? How should an MDS system which is to achieve reasonably good ROUGE score on {\em all} corpora look like?}}
From Table \ref{tab:rfa} within studied systems, in terms of ROUGE-1, ICSISumm achieves the best score on DUC, TextRank on TAC, LexRank on Opinosis, CopyTransformer on Multinews and LexRank on CQASumm.
Hence as of today, no summarization system strictly outperforms others on every corpus. We also see that CopyTransformer which achieves state-of-the-art performance on Multinews achieves 10 points less than the best system on DUC. Similarly, LexRank, the state-of-the-art performer on CQASumm, achieves almost 12 points less than the best system on TAC. Therefore, a system that performs reasonably well across all corpora, is also missing. This is because \textit{different corpora are high on various bias metrics, and summarization systems designed for a particular corpus take advantage and even aggravate these biases}. For example, summarization systems proposed on news based corpora are known to feed only the first few hundred tokens to neural models, thus taking advantage of the layout bias. Feeding entire documents to these networks have shown relatively lower performance. Systems such as LexRank are known to perform well on candidate documents with high inter-document similarity (e.g., Opinosis). Solving the summarization problem for an unbiased corpus is a harder problem, and for a system to be able to perform reasonably well on any test set, it should be optimized to work on such corpora. \\

\begin{table}[!t]
  \centering
  \caption{Pearson correlation between corpus and system with column 4 ({\bf First}) between Abstractness of corpora and system, column 5 ({\bf Second}) between Abstractness of corpora and ROUGE-1 score of systems across datasets and column 6 ({\bf Third})  showing Layout Bias correlation between system and corpora.}
  \scalebox{0.75}{
\begin{tabular}{|l|c|c|l|l|l|}
\hline
\multicolumn{1}{|c|}{}                                  & \multicolumn{5}{c|}{\textbf{Metric}}                                                                                                                                                          \\ \cline{2-6} 
\multicolumn{1}{|c|}{}                                  &                                      &                                     & \multicolumn{3}{c|}{\textbf{Layout correlation}}                                                                 \\ \cline{4-6} 
\multicolumn{1}{|c|}{\multirow{-3}{*}{\textbf{System}}} & \multirow{-2}{*}{\textbf{Abs. corr}} & \multirow{-2}{*}{\textbf{R-1 corr}} & \multicolumn{1}{c|}{\textbf{First}} & \multicolumn{1}{c|}{\textbf{Second}} & \multicolumn{1}{c|}{\textbf{Third}} \\ \hline
LexRank                                                 & -                                    & 0.08                                & \cellcolor[HTML]{EFEFEF}0.88        & \cellcolor[HTML]{EFEFEF}0.06         & \cellcolor[HTML]{EFEFEF}0.96        \\
TextRank                                                & -                                    & -0.24                               & \cellcolor[HTML]{EFEFEF}0.91        & \cellcolor[HTML]{EFEFEF}0.76         & \cellcolor[HTML]{EFEFEF}0.97        \\
MMR                                                     & -                                    & 0.32                                & \cellcolor[HTML]{EFEFEF}0.86        & \cellcolor[HTML]{EFEFEF}0.09         & \cellcolor[HTML]{EFEFEF}0.97        \\
ICSISumm                                                & -                                    & 0.11                                & \cellcolor[HTML]{EFEFEF}0.39        & \cellcolor[HTML]{EFEFEF}0.53         & \cellcolor[HTML]{EFEFEF}0.72        \\
PG                                                      & 0.57                                 & 0.65                                & \cellcolor[HTML]{EFEFEF}0.80        & \cellcolor[HTML]{EFEFEF}-0.80        & \cellcolor[HTML]{EFEFEF}-0.98       \\
PG-MMR                                                  & 0.57                                 & 0.33                                & \cellcolor[HTML]{EFEFEF}0.84        & \cellcolor[HTML]{EFEFEF}-0.69        & \cellcolor[HTML]{EFEFEF}-0.91       \\
CopyTrans.                                              & 0.47                                 & 0.50                                & \cellcolor[HTML]{EFEFEF}0.84        & \cellcolor[HTML]{EFEFEF}-0.31        & \cellcolor[HTML]{EFEFEF}-0.79       \\
Hi - Map                                                & 0.11                                 & 0.45                                & \cellcolor[HTML]{EFEFEF}0.74        & \cellcolor[HTML]{EFEFEF}-0.11        & \cellcolor[HTML]{EFEFEF}-0.46       \\ \hline
\end{tabular}}
\label{tab:correlation}
\end{table}

\noindent\textbf{\textit{ Q3. Why do systems show bias on different metrics, and which other system and corpus attributes are the reason behind it?}}
We begin by studying how {\em abstractness of generated summaries is related to the abstractness of corpora the system is trained on}. For this, we calculate the Pearson correlation coefficient between the abstractness of generated summaries and references across different datasets. From Table \ref{tab:correlation}, we infer that PG, PG-MMR and CopyTransformer show a positive correlation which implies that they are \textit{likely to generate more abstract summaries if the datasets on which they are trained have more abstract references}. 
Lastly, we infer {\em how Layout Bias in system-generated summaries is dependent on the layout bias of reference summaries}. The last three highlighted columns of Table \ref{tab:correlation} infer that the abstractive systems such as PG, PG-MMR, Hi-Map and CopyTransformer show a high negative correlation for the end segments while maintaining a strongly positive one with the starting segment. On the other hand, extractive systems such as LexRank, TextRank, MMR and ICSISumm maintain a strongly positive correlation throughout the segments. On shuffling the source segments internally, we observe that extractive systems tend to retain their correlation with corpora while abstractive systems show no correlation at all (Fig. \ref{fig:position_bias}), proving that in supervised systems, \textit{the layout bias in system summaries propagates from the layout bias present in corpora}.\\

\noindent \textbf{\textit{Q4. Is the task of MDS almost solved, or there is still plenty of scope remaining for improvement?}}
In the previous sections, we computed the oracle extractive upper bound summary using greedy approaches to find the summary that obtains the highest ROUGE score given the candidate documents and references. We observe that the best summarization system on each corpus today obtains a score which is $39.6\%$ of the extractive oracle upper bound on DUC, $47.8\%$ on TAC, $75.02\%$ on Opinosis, $54.5\%$ on Multinews and $49.9\%$ on CQASumm.
This shows that there is enough scope for MDS systems to achieve double the ROUGE scores obtained by the best system to date on each corpus except Opinosis. Therefore, we believe that the task of MDS is only partially solved and considerable efforts need to be devoted to improving the systems. 

\section{Related Work}
Previous attempts to evaluate the quality of the benchmark summarization corpora are few in number and mostly from the time when corpora were manually accumulated. \cite{hirao2004corpus} primarily used the intrinsic metrics of precision and recall to evaluate corpus quality. In addition, the authors proposed an extrinsic metric, called `Pseudo Question Answering'. This metric evaluates whether a summary has an answer to a question that is otherwise answerable by reading the documents or not. Although effective, the cost of such an evaluation is enormous and is not scalable to modern day corpora sizes. For such corpora where multiple references are available, \cite{benikova2016bridging} used an inter-annotator agreement to model the quality of the corpora. They also used non-redundancy, focus, structure, referential clarity, readability, coherence, length, grammaticality, spelling, layout, and overall quality as quantitative features for an MDS corpus. Recently, \cite{ijcai2020-0514} proposed an MDS system that used the baseline PG model along with Hierarchical structural attention to take into account long-term dependencies for superior results compared to baseline models.

There have been a series of very recent studies that look into how to strengthen the definition and discover system biases in single-document summarization. Very recently, \cite{jung2019earlier} studied how position, diversity and importance are significant metrics in analyzing the toughness of single-document summarization corpora. Another recent work \citep{kryscinski2019neural} extensively studied the Layout Bias in news datasets that most single-document summarization systems seem to exploit. Two seminal works, namely \cite{peyrard2019simple} and \cite{peyrard2019studying}, exploited the theoretical complexity of summarization on the ground of importance, analyzing in-depth what makes for a good summary. \cite{peyrard2019simple}  mathematically modeled the previously intuitive concepts of \textit{Redundancy, Relevance and Informativeness} to define \textit{importance} in single-document summarization. \cite{newsroom} proposed a new single-document summarization corpus and quantified how it compares to other datasets in terms of diversity and difficulty of the data. They introduced metrics such as \textit{extractive fragment density} and \textit{extractive fragment coverage} to plot the quality of SDS corpus. {\bf To the best of our knowledge, no comparative work exists for either corpora or systems in MDS, and the current paper is the first in this direction.}
\section{Conclusion}
In this paper, we aimed to study the heterogeneous task of multi-document summarization. We analyzed interactions between widely used corpora and several state-of-the-art systems to arrive at a line of conclusions. We defined MDS as a mapping from a set of non-independent candidate documents to a synopsis that covers \textit{important} and \textit{redundant} content present in the source. We proposed intrinsic metrics to model the quality of an MDS corpus and introduced a framework for future researches to consider while proposing a new corpus. 

We analyzed how  ROUGE ranks of different systems vary differently on different corpora and described what a system that achieves reasonable performance on all corpora would look like. We evaluated how different systems exhibit bias and how their behavior is influenced by corpus properties. We also commented on the future scope for the task of MDS.

Future directions to take forward this work would include a causal analysis of how corpus bias is responsible for bias in model prediction across different corpora and systems. This might bring forward measures to de-bias NLP algorithms with/without de-biasing the corpora.  

\section*{Acknowledgments}
The work was partially supported by the Ramanujan Fellowship and DST (ECR/2017/00l691). T. Chakraborty would like to acknowledge the support of the Infosys Center for AI, IIIT-Delhi.

\bibliographystyle{acl_natbib}
\bibliography{emnlp2020}


\end{document}